\documentclass[11pt,draftcls,onecolumn,twoside]{IEEEtran}
\usepackage{amsmath, amssymb}
\usepackage{multirow}
\usepackage{graphicx}
\usepackage{algorithm}
\usepackage{algorithmic}

\newcommand{\bTheta}{\mbox{\boldmath{$\Theta$}}}

\newcommand{\bnabla}{\mbox{\boldmath{$\nabla$}}}

\newcommand{\bL}{\mbox{\boldmath{$L$}}}

\newcommand{\bU}{\mbox{\boldmath{$U$}}}

\newcommand{\bK}{\mbox{\boldmath{$K$}}}

\newcommand{\bI}{\mbox{\boldmath{$I$}}}
\newcommand{\bv}{\mbox{\boldmath{$v$}}}

\newcommand{\bE}{\mbox{\boldmath{$E$}}}
\newcommand{\bp}{\mbox{\boldmath{$p$}}}

\begin{document}

\title{Connectivity-Enforcing Hough Transform for the Robust Extraction of Line Segments}

\author{Rui F. C. Guerreiro and Pedro M. Q. Aguiar,~\IEEEmembership{Senior Member,~IEEE}
\thanks{The authors are with the Institute for Systems and Robotics / Instituto Superior T\'{e}cnico, Lisboa, Portugal. Contact author: P.~Aguiar, Ph.: +351-21-8418283, Fax: +351-21-8418291, e-mail: aguiar@isr.ist.utl.pt.}}

%
%

\markboth{IEEE Transactions on Image Processing}%
{Guerreiro and Aguiar: Connectivity-Enforcing Hough Transform}

\maketitle

\begin{abstract}
Global voting schemes based on the Hough transform (HT) have been widely used to robustly detect lines in images. However, since the votes do not take line connectivity into account, these methods do not deal well with cluttered images. In opposition, the so-called local methods enforce connectivity but lack robustness to deal with challenging situations that occur in many realistic scenarios, {\it e.g.}, when line segments cross or when long segments are corrupted. In this paper, we address the critical limitations of the HT as a line segment extractor by incorporating connectivity in the voting process. This is done by only accounting for the contributions of edge points lying in increasingly larger neighborhoods and whose position and directional content agree with potential line segments. As a result, our method, which we call STRAIGHT (Segment exTRAction by connectivity-enforcInG HT), extracts the longest connected segments in each location of the image, thus also integrating into the HT voting process the usually separate step of individual segment extraction. The usage of the Hough space mapping and a corresponding hierarchical implementation make our approach computationally feasible. We present experiments that illustrate, with synthetic and real images, how STRAIGHT succeeds in extracting complete segments in several situations where current methods fail.
\end{abstract}

\begin{IEEEkeywords}
\noindent Hough transform, line segment detection, connectivity, connected segments, line pattern analysis, edge analysis.
\end{IEEEkeywords}

 \begin{center} \bfseries EDICS Category: ARS-RBS \end{center}
%
\IEEEpeerreviewmaketitle

\section{Introduction}
\label{sec:intro}

\IEEEPARstart{L}{ine} segments are fundamental low-level features for the analysis of many real-life images. In fact, most man-made objects are made of flat surfaces, originating images with edge maps composed by line segments. Thus,
these segments provide important information about the geometric content of the imaged scene. This has been exploited, {\it e.g.}, for localizing vanishing points \cite{Zhang02videocompass} or to match line segments across distinct views \cite{Schmid97automaticline}. Since more elaborated shapes are often described in an economic way in terms of line segments, their extraction is often a first step in many other problems, {\it e.g.}, rectangle detection~\cite{rectangleDetection08}, the inference of shape from lines~\cite{shapeFromLines96}, map-to-image registration~\cite{MapToImageRegistration}, 3D reconstruction~\cite{lineDrawingTo3D}, or even image compression~\cite{Ageenko98compressionof}.

In this paper, we propose a new method to extract line segments from images in an automatic way. Although this problem has been the focus of attention of several researchers in the past decades, current solutions do not cope with many challenging situations that arise in practice, as we detail in the sequel. For this reason, the robust detection of line segments remains an open frontier (see \cite{Dahyot08pami,borkar2011} for examples of recent advances).

\subsection{Overview of methods for line segment extraction}

The Hough transform (HT)~\cite{Hough62,DudaHart72} is the most popular method to detect lines in images. Basically, the HT extracts the lines that contain larger number of edge points. The key to an efficient implementation is the usage of the Hough space, a two-dimensional space parameterized in such a way that each point in this space represents a line in the image. Each edge point in the image is then mapped to the region of the Hough space that represents the pencil of all the image lines that go through that edge point. By processing all edge points, the votes for each location in the Hough space are accumulated (this space is also referred to as the accumulator array) and the locations with larger number of votes correspond to the parameterizations of the predominant lines in the image. Naturally, the success of the HT comes from its global nature, since all points in a line contribute to its detection.

A  strong limitation of the HT comes from the fact that the voting scheme does not take into account that lines are alignments of {\it connected} points. In fact, since edge points voting for a particular line may be disconnected from each other, there is not guarantee that parameters receiving large numbers of votes correspond to long lines in the image (due to textures and/or noise, a peak in the accumulator array may even correspond to the parameterization of a ``false" line, {\it i.e.}, a line that collects many separate points but is not at all perceived as a line in the image). This problem is particularly critical when using the HT to extract a line {\it segment}, {\it i.e.}, a rectilinear point alignment with length that can be much smaller than the image dimensions. In first place, short segments originate small peaks in the Hough space, which are hard to identify~\cite{FastHoughTransformMore95}; besides, a non-trivial extra step to determine the segment start and end points is required, {\it e.g.}, the analysis of the shape of the spread of votes in the accumulator array~\cite{butterflyHough98}.

\begin{figure}[hbt!]
\centerline{
\includegraphics[height=.225\linewidth]{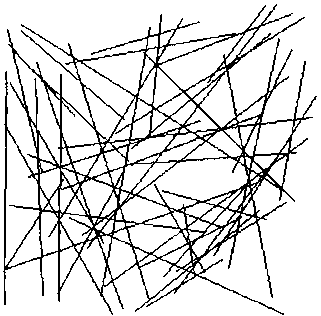}\hspace*{.75cm}
\includegraphics[height=.225\linewidth]{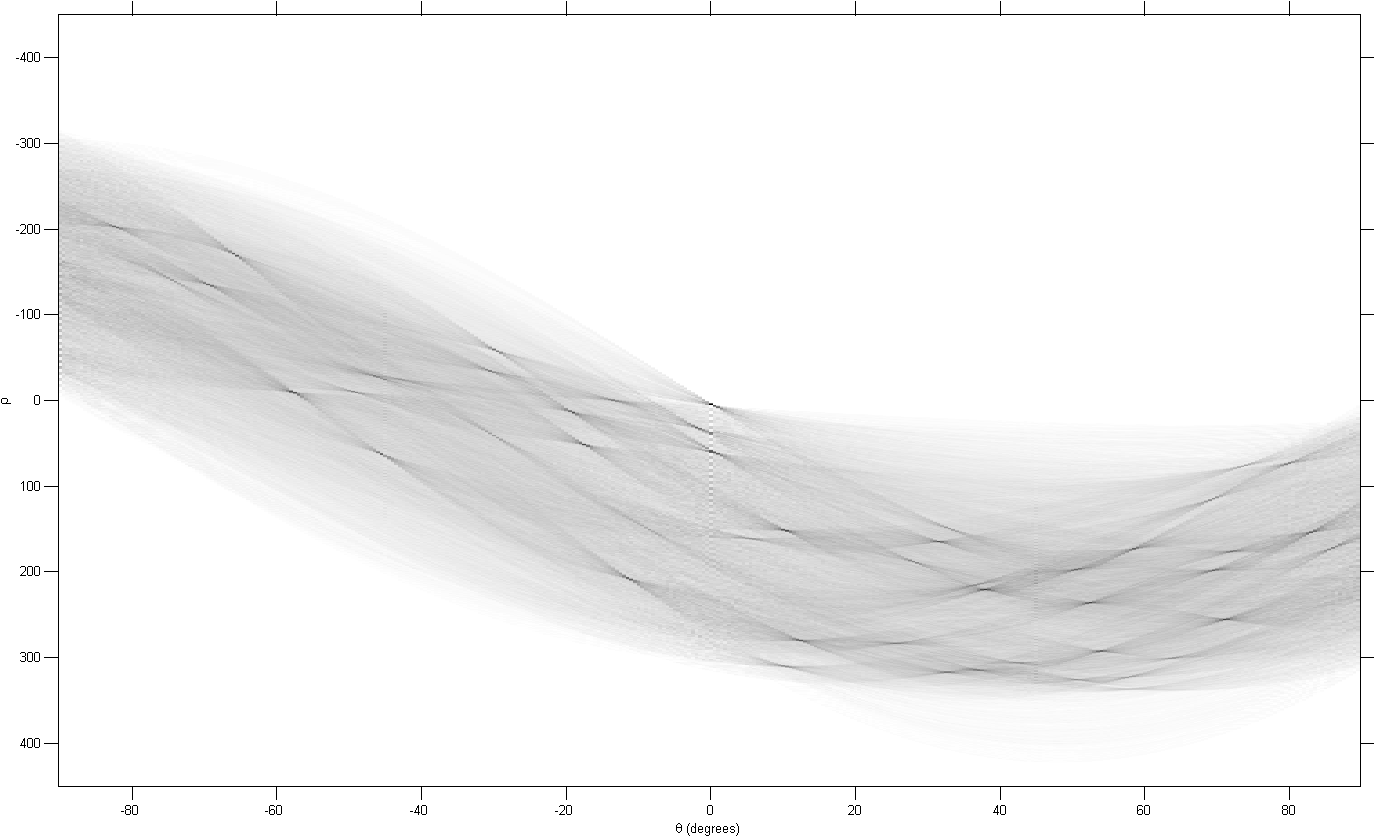}\hspace*{.75cm}
\includegraphics[height=.225\linewidth]{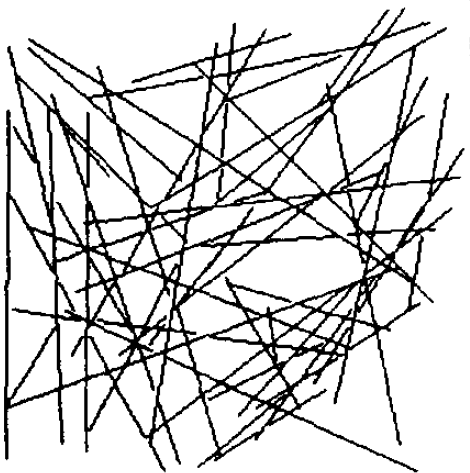}
}\vspace*{.25cm}
\centerline{
\includegraphics[height=.225\linewidth]{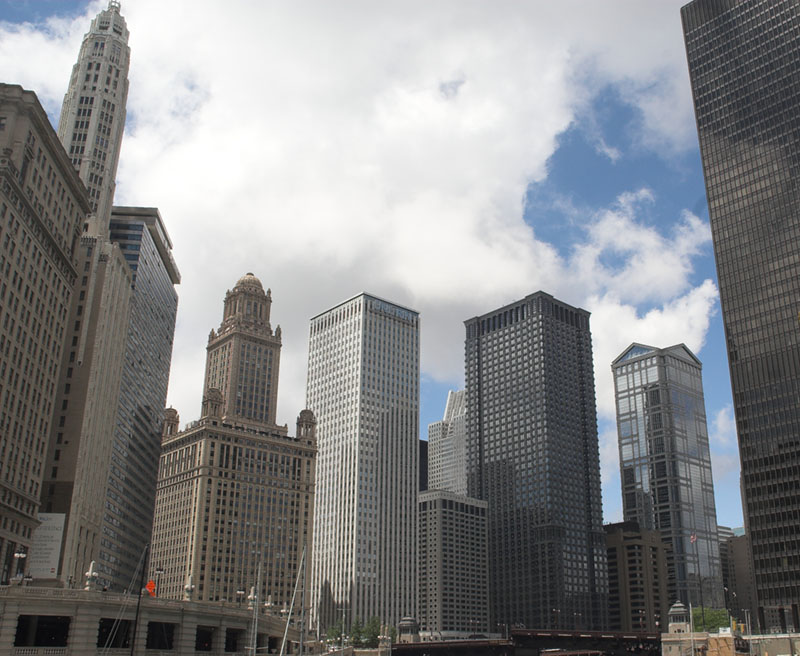}\hspace*{.25cm}
\includegraphics[height=.225\linewidth]{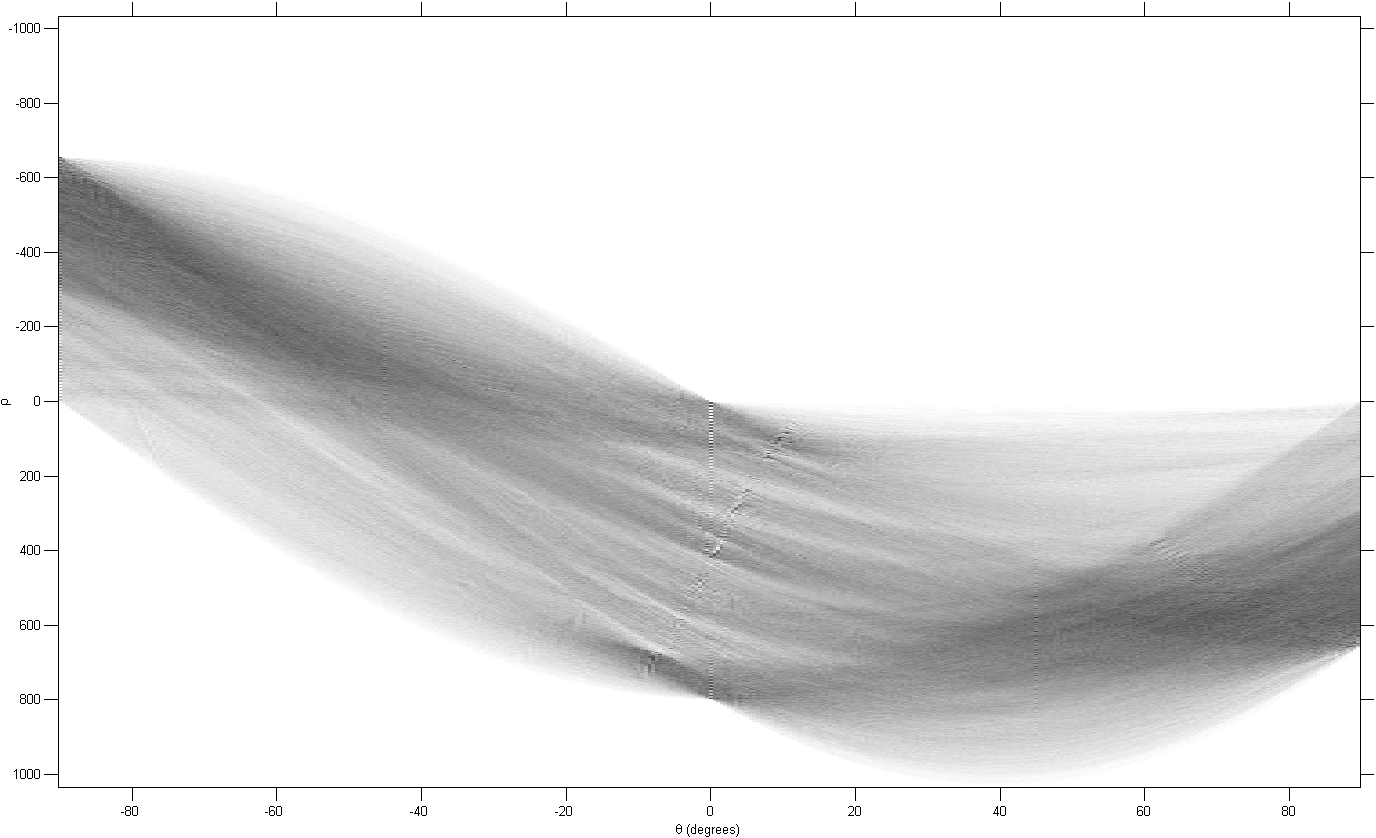}\hspace*{.25cm}
\includegraphics[height=.225\linewidth]{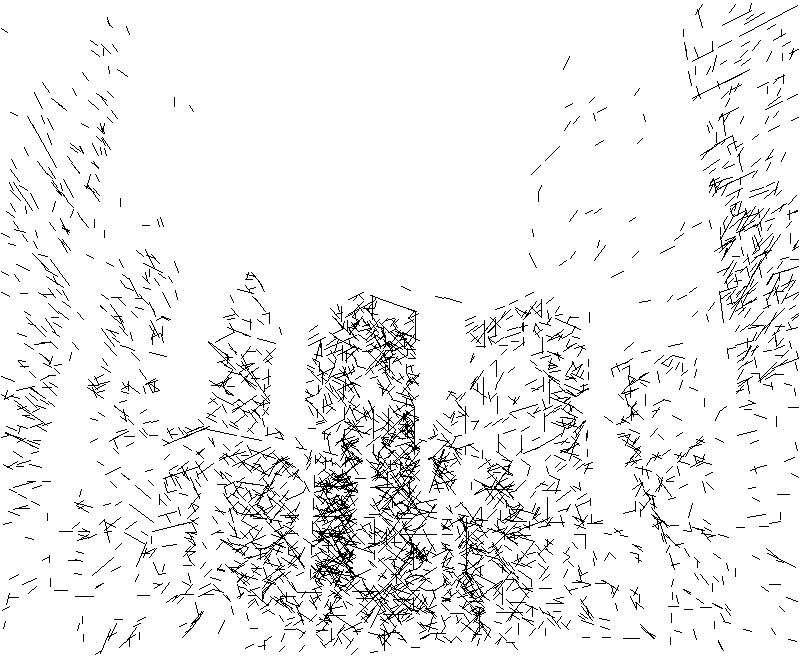}
}
\caption{Top: success of the HT when detecting long lines in a clutter-free image. Bottom: limitation of the HT when detecting short line segments in a complex image. In each case: on the left, the original image, on the middle, the HT accumulator arrays, and on the right, the detected line segments.}\label{fig:houghIllustration}
\end{figure}

Fig.~\ref{fig:houghIllustration} illustrates the difficulty that may arise when using the HT to detect short line segments. We contrast the HT of a synthetic image with long line segments (top left) with the one of a complex real image (bottom left). In the first case, since the number of lines is not very large and they are very long, the HT accumulator array exhibits the expected peaks (darker points visible in the top middle image), which are easily detected, and correspond to the correct lines (top right). In the second case, the large number of edge points in the real image, many of them forming very short segments or only corresponding to texture or noise, originates large numbers of votes for lines that do not correspond to connected segments. As a consequence, the accumulator array does not exhibit prominent peaks (see the smoothness of its grey-level representation in the bottom middle image) and its processing originates a poor result (bottom right) that contains spurious short segments and misses several others, including longer ones.

The limitations of the HT have been pointed out by several authors, {\it e.g.}, \cite{FastHoughTransformMore95,Guru2004,linesPCA06,LSD10} and many efforts have been made to alleviate its problems. For example, the method in \cite{HoughSurvey87} uses the edge direction to reduce the accumulation of spurious votes in the Hough space and \cite{FastHoughTransformMore95} proposes a strategy that is based on processing a line a time: after detecting the line corresponding to the largest peak in the accumulator array, the votes of the edge points that belong to this line are removed. Naturally, the success of this approach hinges on the correctness of the first lines detected. As a consequence, in many practical situations, when facing highly textured images and/or in the presence of noise, these approaches are unable to provide accurate results. In fact, this happens because these improvements do not tackle the fundamental limitation of the HT: the fact that it does not take into account the required connectivity of the edge points forming a line segment. Other authors addressed storage and computational issues of the HT by proposing a hierarchical scheme \cite{FastHoughTransform86}, multiple accumulator resolutions \cite{AdaptiveHoughTransform87}, or the usage of a random sampling of the edge map (probabilistic HT) \cite{probalisticAndNonProbabilistic95}.

In spite of these limitations, the global nature of the HT is attractive, since line parameters estimated from the complete data are naturally more accurate than what can be done locally. However, few papers have approached the problem of developing global methods for line segment extraction, {\it i.e.}, global methods to extract, simultaneously, the line parameters and its extremes. A fruitful example is the method in \cite{LSDAllCombinations06}, which searches among all possible line segment candidates, using a Helmholtz principle for validating. Naturally, the good results come at a high computational cost. Other approaches use the widely known random sample consensus (RANSAC) \cite{Ransac81}: basically, two edge points from the edge map, randomly sampled, define a candidate line; then, the consensus of the line is evaluated by counting the number of other edge points that fit that line segment, given an error tolerance; for segments with a high consensus, the parameters are refined by using an iterative expectation-maximization (EM) method \cite{Hartley2004}. Since the success of RANSAC hinges on the usage of a very large number of samples, these approaches result computationally too complex for many realistic applications.

For the reasons above, the majority of methods for line segment extraction rely on local decisions, rather than on global ones, see \cite{Nevatia1980257,extractingStraightLines86,Guru2004,comparisonLineExtraction07,LSD10} for examples. These local methods outperform the HT by taking (local) connectivity into account, and result computationally simple, but lack robustness to deal with challenging situations, {\it e.g.}, when line segments cross. Furthermore, their local nature make long line segments particularly difficult to extract in many realistic scenarios, because, due to noise and clutter, these segments are interrupted. The majority of local methods use three steps: first, obtaining a region of connected edge points; then, roughly estimating the line segment direction; and finally, refining and extending the segment by including new edge points that approximately fit the line.
The first step consists of chaining edge points \cite{edgeChaining92}. Methods such as the one in \cite{edgeChainLineCutting92} even skip the subsequent line fitting and refinement steps by chaining connected edge points into curves and then cutting them into line segments, using a straightness criterion. Texture, low-contrast regions, crossing segments, and noise make difficult the extraction of large connected regions belonging to a single segment.
The second step consists of fitting a line to the chain of edge points using, {\it e.g.}, total least-squares (TLS) \cite{comparisonLineExtraction07}. Naturally, the reliability of the regression depends on the length of the underlying point chain. Some methods bypass the chaining of edge points: \cite{featureExtraction78} uses the so-called local HT (LHT) \cite{twoFrame03}, roughly estimating the segment direction from the peaks of local orientation histograms, computed at each edge point; \cite{Guru2004,directRegression97} directly fit a line to all edge points inside a sliding window, which only provides reasonable estimates for simple scenes, with very small clutter.
The final step usually involves alternating between two stages until convergence \cite{comparisonLineExtraction07}: inclusion of new edge points that are close to the candidate line, according to a distance measure; and re-estimation of the line segment parameters from the new set of edge points. As it is typical with this type of methods, a poor initial model for the line segment model may compromise the final result. Furthermore, the process may terminate too early when attempting to extract a long line segment, due to the common cluttered nature of the edge maps of real images.
Two popular local methods for line segment detection are \cite{extractingStraightLines86} and the LSD (Line Segment Detector) of \cite{LSD10}. The method in \cite{extractingStraightLines86} coarsely quantizes the local orientation angles, chains adjacent pixels with identical orientation labels, and fits a line segment to the grouped pixels. LSD \cite{LSD10} extends this idea by using continuous angles and eliminates false line segment detections by using the Helmholtz principle of \cite{LSDAllCombinations06}.

\subsection{Proposed approach}

The key ingredient of our method is the incorporation of connectivity into the HT voting process. This is done by imposing that edge points only vote for lines according to which they are spatially connected to other points. As a consequence, the vast majority of spurious votes are eliminated and peaks in the accumulator array become prominent and truly correspondent to line segments of maximum length. Simultaneously, our method, which we call STRAIGHT (Segment exTRAction by connectivity-enforcInG Hough Transform), integrates into the voting process the usually separate step of determining the extremes.

STRAIGHT starts by computing the prominent directions at each edge point, which will guide the search for the orientations of line segments. An image line segment is characterized by a rectilinear alignment of dark-to-light (or opposite) transitions. Our prominent direction detector computes, for each edge point, the set of directions according to which there is a predominance of those intensity transitions. This is accomplished by extending the  LHT~\cite{twoFrame03} to only take into account directionally coherent edge points: in a first step, signed directional edge maps are computed; then, orientation histograms are built at each edge point, by considering the neighboring edge points whose relative positions agree with the angle of the directional edge map. The histogram accumulates the signed values of the intensity transitions, thus the prominent directions at each edge point are detected by finding large magnitude entries in the corresponding histogram.

After computing the local prominent directions, STRAIGHT extracts line segments using the knowledge that the edge points forming each of them must be connected. This is done by computing new LHT-like maps (which we will call {\it length maps}) for each edge point, this time taking into account all other edge points whose position and directional content agree with potential line segments. Position matters because only points that respect connectivity are considered; directional content matters because only edge points with prominent direction that agrees with the candidate segment are considered. In practice, for each prominent direction of each edge point, STRAIGHT progressively considers edge points further away until the connectivity criterion is violated. After exploring all candidate directions, the ones that collected more distant edge points correspond to the orientations of the longest connected line segments going through the starting point. Note that allowing a set of prominent directions, rather than a single one, enables dealing with crossing segments.

Our implementation of STRAIGHT incorporates the explicit mapping of uncertainty balls around the edge points into the Hough space, increasing robustness and accuracy, and uses a hierarchical coarse-to-fine strategy to explore candidate directions, leading to a computationally tractable algorithm. We present illustrative results of experiments that use synthetic and real images to compare STRAIGHT with the HT \cite{DudaHart72} and the state-of-the-art local method LSD \cite{LSD10}. A preliminary version of this work is in \cite{guerreiro-icip11}.

\subsection{Paper organization}

The organization of the remaining of the paper is as follows. Section~\ref{sec:section2} describes the computation of local prominent directions. Section~\ref{sec:section3} details the extraction of line segments by enforcing connectivity. The hierarchical implementation is described in Section \ref{sec:section4}. Experimental results are reported in Section~\ref{sec:section5} and Section~\ref{sec:conclusions} concludes the paper.

\section{Computing Local Prominent Directions}
\label{sec:section2}

In terms of image intensities, the existence of a line segment corresponds to the presence of a rectilinear alignment of dark-to-light (or opposite) transitions. Despite the multiple sources of inaccuracies in edge detection, such as noise and clutter, there should be a predominance of either positive or negative intensity variations (corresponding to light-to-dark and dark-to-light transitions, respectively) along the line segment. This predominance is what we exploit to compute the prominent directions at each edge point, {\it i.e.}, the set of possible orientations of line segments going trough that point.

We capture the local directional content of an image $\bI$ by computing its derivatives, through the convolution with four oriented kernels,
\begin{equation}
\bnabla_{\!\theta} \bI= \bI * \bK_{\!\theta}\,,\quad\theta\in\{0^{\circ}, 45^{\circ}, 90^{\circ}, 135^{\circ}\}\,.\label{eq:og}
\end{equation}
Although kernels with a large support would smooth the noise, we use simple standard central difference kernels, since they enable more precise edge localization and angular responses, by minimizing the influence of surrounding pixels. Thus, we set
\[
\bK_{0}=\!\left[
          \begin{array}{rrr}
            0 & 0 & 0 \\
            1 & 0 & -1 \\
            0 & 0 & 0
          \end{array}
        \right]\!,\;
\bK_{45}=\!\left[
          \begin{array}{rrr}
            0 & 0 & -1 \\
            0 & 0 & 0 \\
            1 & 0 & 0
          \end{array}
        \right]\!,\;
\bK_{90}=\!\left[
          \begin{array}{rrr}
            0 & -1 & 0 \\
            0 & 0 & 0 \\
            0 & 1 & 0
          \end{array}
        \right]\!,\;
\bK_{135}=\!\left[
          \begin{array}{rrr}
            -1 & 0 & 0 \\
            0 & 0 & 0 \\
            0 & 0 & 1
          \end{array}
        \right]\!.
\]
In our case, robustness to noise when extracting line segments comes from the requirement of point connectivity, as detailed in the following section.

The directional edge maps $\bE_{\theta}$, $\theta \in \{0^{\circ}, 45^{\circ}, 90^{\circ}, 135^{\circ}\}$, are obtained by thresholding the derivatives and retaining their sign, {\it i.e.},
\[
\bE_{\!\theta} (x,y)= \left\{\begin{array}{rcl}1 & & \text{if} \;\; \bnabla_{\!\theta}\bI(x,y) \geq T \\ -1 & & \text{if} \;\; \bnabla_{\!\theta}\bI(x,y) \leq -T \\ 0 & & \text{otherwise}\,.\end{array}\right.
\]
We call $(x,y)$ an edge point when $|\bE_{\!\theta} (x,y)| = 1$ for at least one value of $\theta$.

\begin{figure}[htb]
\centerline{\includegraphics[width=0.3\linewidth]{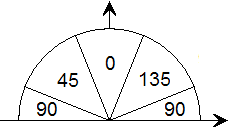}}
\caption{Selection of directional edge map $\bE_{\!\theta}$ in terms of the relative position between edge points.}\label{fig:edgeQuadrants}
\end{figure}

We extend the LHT \cite{twoFrame03} to take into account the edge directional content, {\it i.e.}, our method builds local orientation histograms by using the neighboring edge points whose direction is coherent with its position. To clarify, when building the histogram for the edge point $(x_0,y_0)$, we first compute the directions of the segments passing through $(x_0,y_0)$ and each neighboring edge point $(x,y)$,
\[
\theta_{(x_0,y_0)}\left(x,y\right) = \arctan\left(\frac{y-y_0}{x-x_0}\right)\,\in\,[0^{\circ}, 180^{\circ})\,.
\]
Then, for each neighbor $(x,y)$, the histogram count uses only the directional edge map $\bE_{\!\theta}$ with the value of $\theta$ that is closer to direction perpendicular to $\theta_{(x_0,y_0)}\left(x,y\right)$, {\it i.e.},
\begin{equation}\label{eq:computingQuarter}
\begin{array}{rcl}
\bE_{90}(x,y) & \text{if} &  \theta_{(x_0,y_0)}\left(x,y\right) \in [0, 22.5] \cup (155.5, 180)\,, \\
\bE_{135}(x,y) & \text{if} &  \theta_{(x_0,y_0)}\left(x,y\right) \in (22.5, 67.5]\,, \\
\bE_{0}(x,y) & \text{if} & \theta_{(x_0,y_0)}\left(x,y\right) \in (67.5, 112.5]\,, \\
\bE_{45}(x,y) & \text{if} & \theta_{(x_0,y_0)}\left(x,y\right) \in (112.5, 155.5]\,,
\end{array}
\end{equation}
which just corresponds to the representation in Fig.~\ref{fig:edgeQuadrants}. For example, if $(x_0,y_0)=(0,0)$ and $(x,y)=(0,3)$, we have $\theta_{(0,0)}(0,3)=90^{\circ}$, a vertical segment, and the directional edge point that contributes to the histogram is $\bE_{0}(0,3)$, since $\bK_0$ is the kernel that best responds to the horizontal transitions that define vertical edges.

As usual in the LHT, the number $B$ of histogram bins is fixed ($B=32$ is typical) and each edge point contributes to the two bins whose centers approximate the angle $\theta_{(x_0,y_0)}\left(x,y\right)$ by excess and default (the contributions are weighted according to the distance to the bin centers). The signs in the directional edge maps are taken into account through positive or negative contributions to the histogram bins. This way, we filter out conflicting contributions that may occur due to noise, textures and image clutter. We use a relatively large neighborhood for the local histograms (a $7$-pixel radius circular window) to minimize the influence of the spurious points in the edge maps. The prominent directions at each edge point are found by thresholding the magnitude of the corresponding local histogram (we use the threshold of $50 \%$ of the number of edge points inside the circular window for each direction, thus a direction is considered prominent when the majority of the edge points in that direction have the same intensity transition sign).

To represent the range of possible directions going through each edge point $(x_0,y_0)$, we store the set of prominent bin centers and the corresponding image gradients, {\it i.e.},
\[
\bTheta(x_0,y_0) = \left\{\left({\theta_1},\bnabla_{\!\theta_1} \bI(x_0,y_0)\right), \left({\theta_2},\bnabla_{\!\theta_2} \bI(x_0,y_0)\right),  \hdots, \left({\theta_N},\bnabla_{\!\theta_N} \bI(x_0,y_0)\right)\right\}\,,
\]
where $0\leq N\leq B$ is the number of histogram bins whose count was above the threshold. If $N\geq 1$, for $1\leq n\leq N$, we denote by ${\theta_n}$ the central angle of the bin $n$ and by $\bnabla_{\!\theta_n} \bI(x_0,y_0)$ the image gradient \eqref{eq:og}, with orientation $\theta$ that best matches ${\theta_n}$, according to \eqref{eq:computingQuarter}. To account for the histogram discretization, {\it i.e.}, the nonzero width of the bins, we consider in the sequel as possible directions of line segments all the orientations $\theta\in[\theta_n-\Delta_{\theta},\theta_n+\Delta_{\theta}]$, with $\Delta_{\theta}=180/B/2=90/B$.

\section{Extracting Connected Line Segments}
\label{sec:section3}

We now describe the core of STRAIGHT, {\it i.e.}, the way we incorporate connectivity into the line segment extraction. We start by making explicit the parameter search problem that underlies the extraction of each of the segments, introducing the {\it length map}, which plays a similar role of the HT accumulator array. Then, we describe how edge points are sequentially mapped into the length map by taking into account both the edge point connectivity and the uncertainty due to discretization. Finally, we describe the procedure to extract line segments from the length map.

\subsection{Line segment extraction as a parameter search problem -- the length map}

\begin{figure}[htb]
\begin{minipage}[b]{\linewidth}
\centering
 \centerline{\includegraphics[width=0.3\linewidth]{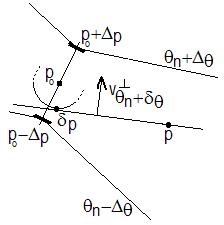}}
\end{minipage}
\caption{Edge point $\bp_0$, prominent direction $\theta_{n}$, line segment specified by parameters $(\delta_p,\delta_{\theta})$, and range limits $\Delta_p$ and $\Delta_\theta$.}\label{fig:aLine}
\end{figure}

Let $\bp_0=(x_0,y_0)$ be an edge point and $\bTheta(\bp_0)$ the set of prominent directions of possible line segments passing through $\bp_0$, as introduced in the previous section. To accurately detect a line segment in the image it is necessary to estimate the sub-pixel location of the line, which will be parameterized by its position and orientation, in a similar way to the HT. The line position is specified in terms of its distance $\delta_p$ to the edge point $\bp_0$. The line orientation is represented by $\delta_{\theta}$, which represents the deviation with respect to the prominent direction angle $\theta_{n}$ in $\bTheta(\bp_0)$. Thus, as illustrated in Fig.~\ref{fig:aLine}, any point $\bp=(x,y)$ belonging to the line $(\delta_p,\delta_{\theta})$ obeys
\begin{align}
\langle\bp,\bv_{\theta_n+\delta_{\theta}}^{\bot}\rangle = \langle\bp_0,\bv_{\theta_n+\delta_{\theta}}^{\bot}\rangle + \delta_p\,,\label{eq:basicLine}
\end{align}
where $\langle\cdot,\cdot\rangle$ is the standard inner product and $\bv_{\theta}^{\bot} = (\sin(\theta),-\cos(\theta))$ is a unit vector with directional orthogonal to $\theta$. The candidate line segment is allowed to deviate at most a predefined $\Delta_p$ from $\bp_0$, {\it i.e.}, $\delta_p \in [-\Delta_p, \Delta_p]$, (in our experiments, we use $\Delta_p=1$) and $\Delta_\theta$ from $\theta_n$, {\it i.e.}, $\delta_\theta \in [-\Delta_\theta, \Delta_\theta]$ ($\Delta_\theta$ was defined in the previous section).

When the estimate $(\delta_p,\delta_{\theta})$, coincides with the true value of the parameters defining a line segment in the image, there are several
other edge points along the line $(\delta_p,\delta_{\theta})$ for which the orientation $\theta = \theta_n + \delta_{\theta}$ is also prominent (with matching signs of the image gradient), according to the information collected in $\left\{\bTheta(\cdot)\right\}$. We call a {\it candidate match} to each of these edge points. Besides, due to the connectivity of the edge points forming a line, the gap between candidate matches must be smaller than a predefined {\it maximum distance threshold} $d$. Naturally, the closer the estimated line is to the actual one, more distant edge points of the true line segment are captured. This is the key point of our approach, which formalizes the extraction of a line segment as the search for the parameters $(\delta_p,\delta_{\theta})$ that {\it maximize the length} of the segment that can be extracted in the neighborhood of $\bp_0$.

To extract the maximum length segments that pass close to each edge point, we borrow inspiration in the LHT, where local accumulator arrays are used in contrast to the single accumulator array of the HT, which can not discriminate between distinct segments falling in the same line. In our case, we define local 2D {\it length maps} $\bL: [-\Delta_p, \Delta_p]\times  [-\Delta_{\theta}, \Delta_{\theta}] \mapsto \mathbb{N}_0$. For each edge point, $\bL(\delta_p, \delta_{\theta})$ will contain the integer length of the line segment $(\delta_p, \delta_{\theta})$. This map can be regarded as an extension of the accumulator array of a LHT in order to take line segment connectivity into account. In this scenario, the extraction of a line segment passing close to $\bp_0$ consists of filling $\bL(\cdot,\cdot)$, obtaining the parameters $(\delta_{p},\delta_{\theta})$ for which $\bL(\delta_p,\delta_{\theta})$ is maximum, and computing its start and end points.

To fill $\bL(\cdot,\cdot)$ through the direct implementation of an exhaustive scanning of the space $[-\Delta_p, \Delta_p]\times  [-\Delta_{\theta}, \Delta_{\theta}]$ would be computationally unbearable. In fact, the discretization of this space must be very fine to yield accurate results and, for each location, many edge points have to be processed (possibly, hundreds or thousands). Besides, this approach would use redundant computations, since each edge point would be visited several times because it may belong to several candidate line segments. For this reason, as usually done to fill the HT accumulator array, we also adopt a pixel-centered approach, where the edge points are used to fill the length map $\bL(\cdot,\cdot)$ in an efficient way.

\subsection{Incorporating uncertainty -- the update region}

\begin{figure}[htb]
 \centerline{
\includegraphics[width=0.3\linewidth]{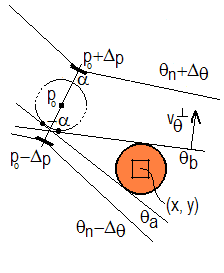}\hspace*{.75cm}
\includegraphics[width=0.08\linewidth]{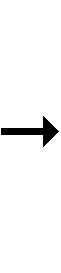}\hspace*{.25cm}
\includegraphics[width=0.4\linewidth]{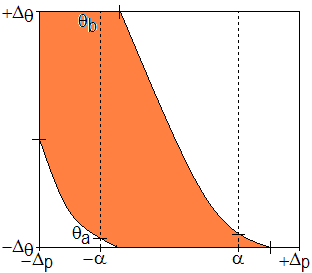}
}
\caption{The uncertainty ball of an edge pixel (left) and the corresponding update region in the length map domain (right).}\label{fig:rangeOfUpdates}
\end{figure}

We now show how each individual edge point is processed in our pixel-centered approach. Due to the pixel grid discretization, we model each edge point by an uncertainty ball, rather than a pointwise feature. We use the uncertainty ball radius $R = 1$, the maximum expected error in the location of each pixel. Because the uncertainty ball has a non-infinitesimal area, there is a set of parameters $\{(\delta_p, \delta_{\theta})\}$, whose corresponding line segments cross it. This is illustrated in the left side of Fig.~\ref{fig:rangeOfUpdates}, where the lines formed by all orientations between $\theta_a$ and $\theta_b$ cross the uncertainty ball centered at $(x,y)$, for position deviation $\delta_p = -\alpha$ from $\bp_0$. We call {\it update region} of the length map domain to the set of positions and orientations $\{(\delta_p, \delta_{\theta})\}$, whose corresponding line segments cross the uncertainty ball centered at each edge pixel. The update region for the pixel $\bp=(x,y)$, in the length map of $\bp_0$, is shown on the right side of Fig.~\ref{fig:rangeOfUpdates}.

To find the analytic expressions for the bounds of update region, we use the line equation \eqref{eq:basicLine}, now seen as a condition for line segments rather than points. When computing the length map for the edge point $\bp_0$, we see from \eqref{eq:basicLine} that any segment $(\delta_p,\delta_{\theta})$ that crosses the uncertainty ball of pixel $\bp=(x,y)$ must verify
\begin{align}
\langle\bp,\bv_{\theta_n+\delta_{\theta}}^{\bot}\rangle = \langle\bp_0,\bv_{\theta_n+\delta_{\theta}}^{\bot}\rangle + \left(\delta_p-r\right)\,,\label{eq:basicLine23}
\end{align}
where $-R\leq r\leq R$ is the distance between the center of the ball and the segment, along vector $\bv_{\theta_n+\delta_{\theta}}^{\bot}$ (depicted in Fig.~\ref{fig:rangeOfUpdates}). From \eqref{eq:basicLine23}, the line segment position $\delta_p$ is easily expressed in terms of its orientation $\delta_{\theta}$ and $r$:
\begin{equation}
\delta_p =\langle\bp-\bp_0,\bv_{\theta_n+\delta_{\theta}}^{\bot}\rangle+r\,.\label{eq:deltap}
\end{equation}
The update region, which we denote by $\bU$, is thus the collection of intervals specified by all possible orientations $\delta_{\theta}$ and corresponding positions $\delta_p$ given by \eqref{eq:deltap}, with $|r|\leq R$:
\begin{align}
\bU = \left\{(\delta_p,\delta_{\theta}) : \delta_{\theta} \in \left[-\Delta_{\theta}, \Delta_{\theta}\right] , \delta_p \in \left[\delta_p^{-}(\delta_{\theta}),\delta_p^{+}(\delta_{\theta})\right] \cap \left[-\Delta_p, \Delta_p\right] \right\}\,,\label{eq:updateLenghtArray}
\end{align}
where the limits $\delta_p^{-}(\delta_{\theta})$ and $\delta_p^{+}(\delta_{\theta})$ are made explicit from \eqref{eq:deltap} by expanding $\bv_{\theta_n+\delta_{\theta}}^{\bot}$:
\begin{equation}
\delta_p^{-}(\delta_{\theta}) = (x-x_0)\sin\left({\theta_n}+\delta_{\theta}\right)
-(y-y_0)\cos\left({\theta_n}+\delta_{\theta}\right)-R\,,\label{eq:pointRanges}
\end{equation}
\begin{equation}
\delta_p^{+}(\delta_{\theta}) = (x-x_0)\sin\left({\theta_n}+\delta_{\theta}\right)
-(y-y_0)\cos\left({\theta_n}+\delta_{\theta}\right)+R\,.
\label{eq:pointRangesII}
\end{equation}
If the update region of a given pixel is non-empty, we say that the pixel is within the {\it search range}. In the illustration of Fig.~\ref{fig:rangeOfUpdates}, the search range for $\bp_0$ is the one limited by the lines labelled with $\theta_n-\Delta_{\theta}$ and $\theta_n+\Delta_{\theta}$.

As geometrically evident, the range of angles of lines that cross an uncertainty ball decreases as the distance between $\bp_0$ and $\bp$ increases (an approximate expression for this range is $2\arcsin\left({R}/{\left\|\bp-\bp_0\right\|}\right)$, obtained by noting that $\overline{\bp_0,\bp}$ can be approximated by the hypotenuse of a right-angled triangle of which $R$ is the small cathetus). As a consequence, to enable the extraction of long line segments, {\it i.e.}, containing edge points $\bp$ far from $\bp_0$, the length map must be densely discretized to sample all relevant values of $\delta_{\theta}$. We propose an efficient way to deal with this need through the hierarchical coarse-to-fine procedure described in the Section \ref{sec:section4}.

We observe that expressions \eqref{eq:pointRanges} and \eqref{eq:pointRangesII} are similar to the parameterizing equation of the HT \cite{DudaHart72}, $\rho = x\cos(\theta)+y\sin(\theta)$, with $\bp_0 = (x_0,y_0)$ as the origin of the coordinate system and $\theta$ shifted by $90^{\circ}$. Thus, the boundaries of the update region resemble the sinusoidal shape of the bundles of votes of each edge point in the HT accumulator array (see Fig.~\ref{fig:rangeOfUpdates} where, in fact, only a segment of that shape is seen, due to the length map limits). In what respects to the resolution of the accumulator array, when using the HT, the contradictory requirements of accuracy (high resolution) and coping with discretization error (low resolution, so that votes of the same line fall within the same bin) makes difficult, if not impossible, to achieve a good compromise. Strategies that uniformly blur the accumulation array ({\it e.g.}, by using multiple resolutions \cite{FastHoughTransform86,AdaptiveHoughTransform87}, or kernels of various sizes \cite{Dahyot08pami}) do not change the scenario, since they still neglect the distinct influence of the discretization error of edge points located at different positions. In opposition, in our case, the resolution of the length map can be chosen arbitrarily large, since we model the actual discretization error of each individual edge point by using the correspondent (position-dependent) update region, as descibed above.

\subsection{Sequential mapping of edge points to the length map}

After describing how each pixel maps to a corresponding update region in the length map, we now show how to fill this map in a sequential way by processing all image edge points. As before, let us consider the case the of the edge point $\bp_0$, with prominent direction $\theta_n$. When filling the corresponding length map $\bL$, we consider the image divided in two half-planes by the line orthogonal to $\theta_n$ passing through $\bp_0$. In each half-plane, starting from $\bp_0$ we circularly scan the image, with progressively larger radius, mapping to the corresponding half-plane length map the candidate matches that fall within the search range and do not violate the connectivity requirement.

\begin{figure}[htb]
 \centerline{\includegraphics[width=0.5\linewidth]{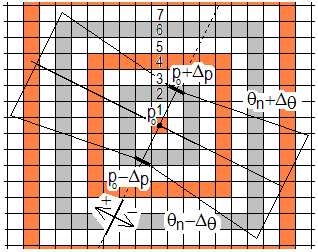}}
\caption{Central edge point $\bp_0$, search range, and equidistant curves, each labeled by its integer distance to $\bp_0$.}\label{fig:equidistantPixels}
\end{figure}

Due to the graceful adaptation to the discrete pixel grid, we use the so-called Manhattan distance to define the {\it equidistant curves} as the set of pixels $\bp$ located at fixed distance $e$ from $\bp_0$ ({\it i.e.}, such that $\| \bp - \bp_0\|_{\infty} = e$). Fig.~\ref{fig:equidistantPixels} illustrates the scenario, with the central edge point $\bp_0$, the equidistant curves, labeled by the distance values, the search range and the half-planes.

\begin{figure}[htb]
 \centerline{\includegraphics[width=0.4\linewidth]{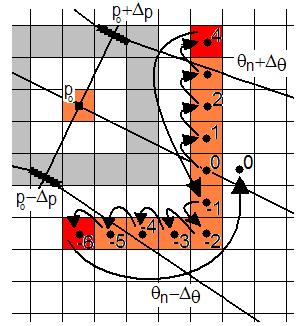}}
\caption{Illustration of the scanning pattern for a single equidistant curve in one of the half planes.}\label{fig:scanning}
\end{figure}

For each half-plane, we scan each equidistant curve, starting with the one closer to $\bp_0$ ({\it i.e.}, the curve with label $e = 1$ in Fig.~\ref{fig:equidistantPixels}), looking for candidate matches. Fig.~\ref{fig:scanning} illustrates the scanning pattern for each equidistant curve. It starts in the center of the search range, {\it i.e.}, the pixel labeled with $0$ in Fig.~\ref{fig:scanning}, and processes each pixel within the curve until reaching the limit of the search range ({\it i.e.}, the pixels labeled with positive values in Fig.~\ref{fig:scanning}, up to label $4$, shown in red). Then, the pixels in the other direction are scanned, until the complete equidistant curve within the search range was processed ({\it i.e.}, the pixels labeled with negative values in Fig.~\ref{fig:scanning}, down to label $-6$, shown in red). Then, the following equidistant curve is processed.

To account for the usage of the Manhattan distance, the discrete pixel $[\bp]$ at the center of the search range for the equidistant curve $e$ is given by
\[[p]=\mbox{round}\left(\bp_0\pm e\frac{\bv_{\theta_n}}{\|\bv_{\theta_n}\|_{\infty}}\right)\,,\]
where $\bv_{\theta} = (\cos(\theta),\sin(\theta))$ is a unit vector with angle $\theta$, and the signal $\pm$ depends on the half-plane being considered.

When a candidate match is found in the equidistant curve $e$, its update region is computed, according to  \eqref{eq:updateLenghtArray}, \eqref{eq:pointRanges}, and \eqref{eq:pointRangesII}, and the corresponding entries of the length map are updated. This updating consists simply in setting those entries to the value of the equidistant curve number, $e$. This indicates that there are valid line segments of (at least) size $e$ with the parameters corresponding to those entries. To capture the connected nature of the line segments, we first prune the update region, eliminating the locations where the difference between the current value of the length map, $\bL$, and the equidistance value $e$ is larger than the maximum distance threshold $d$, {\it i.e.},
\begin{align}
\bU \leftarrow \bU \backslash \left\{ \left[e - \bL(\delta_p,\delta_{\theta})\right]  > d,  \delta_p \in \left[-\Delta_p, \Delta_p\right],\delta_{\theta} \in \left[-\Delta_{\theta}, \Delta_{\theta}\right]\right\}\,.\label{eq:eliminateBadPoints}
\end{align}
Then, we update the length map according to
\[
\bL \leftarrow e\bU + \bL \odot \overline{\bU}\,,
\]
where $\bU$ is seen as a binary mask and $\odot$ denotes the Hadamard, or elementwise, product. When the distance between $e$ and all the values in the length map, $\bL(\cdot,\cdot)$, is larger than $d$, {\it i.e.}, when $\bU = \emptyset$, there are not updatable entries in the length map and the scanning stops for the corresponding half-plane. In the vast majority of our experiments, we used $d = 2$ pixels.

Alg.~\ref{alg:filling} synthesizes the procedure just described to compute each half-plane length map. The final length map for each edge point $\bp_0$ and prominent direction $\theta_n$ is obtained by adding the two half-plane length maps.

\begin{algorithm}[hbt]
   \caption{Filling one half-plane length map $\bL$ for edge point $\bp_0$ and prominent direction $\theta_n$.\label{alg:filling}}
\begin{algorithmic}[1]
\STATE {\bf input:} $\bp_0=(x_0,y_0)$, $\theta_n$, prominent directions $\left\{\bTheta(\cdot)\right\}$, maximum distance $d$, uncertainty radius $R$
\STATE $\bL(\cdot,\cdot)=0$, $e=1$
\REPEAT
    \FOR{$[\bp]=(x,y)$ in the equidistant curve $e$ (as illustrated in Fig.~\ref{fig:scanning})}
    \IF{$[\bp]$ is a {\it candidate match} (according to $\bTheta(\bp)$)}
    \STATE $\bU=\emptyset$
    \FOR{$\delta_{\theta} \in \left[-\Delta_{\theta}, \Delta_{\theta}\right]$}
    \STATE $\delta_p^{-}=(x-x_0)\sin\left({\theta_n}+\delta_{\theta}\right)
-(y-y_0)\cos\left({\theta_n}+\delta_{\theta}\right)-R$
    \STATE $\delta_p^{+}=(x-x_0)\sin\left({\theta_n}+\delta_{\theta}\right)
-(y-y_0)\cos\left({\theta_n}+\delta_{\theta}\right)+R$
    \STATE $\bU \leftarrow \bU \cup \left\{(\delta_p,\delta_{\theta}):\delta_p \in \left[\delta_p^{-},\delta_p^{+}\right] \cap \left[-\Delta_p, \Delta_p\right]\right\}$
    \ENDFOR
    \STATE $\bU \leftarrow \bU \backslash \left\{ e - \bL(\cdot,\cdot)  > d\right\}$ (requirement of line segment connectivity)
    \STATE $\bL \leftarrow e\bU + \bL \odot \overline{\bU}$
    \ENDIF
    \ENDFOR
    \STATE $e\leftarrow e+1$
\UNTIL $e-\max\{\bL(\cdot,\cdot)\}>d$
\STATE {\bf output:} $\bL$
\end{algorithmic}
\end{algorithm}

\subsection{Extracting line segments}

As when detecting lines from the peaks of the HT accumulator array, we detect line segments passing through $\bp_0$ with an orientation close to the prominent direction $\theta_n$ by simply collecting position-orientation pairs lying in the range $(\delta_p, \delta_{\theta})\in [-\Delta_p,\Delta_p]\times [-\Delta_{\theta},\Delta_{\theta}]$ that correspond to peaks in the corresponding length map $\bL(\cdot,\cdot)$.

Whenever a line segment is detected, with position-orientation parameters $(\delta_p,\delta_{\theta})$, we also obtain in a straightforward way the coordinates of its extremes:
\begin{align}
[p_{\pm}]=\mbox{round}\left(\bp_0+E_{\pm} \frac{\bv_{\theta_n+\delta_{\theta}}}{\|\bv_{\theta_n+\delta_{\theta}}\|_{\infty}} +\delta_p \bv_{\theta_n+\delta_{\theta}}^{\bot}\right)\,,\label{eq:lineExtremes}
\end{align}
where the subscript $\pm$ differentiates both extremes and $E_{\pm}$ denotes the maximum values of the length map of the corresponding half-planes.

To prevent multiple detections of a single line segment, each time a segment is detected for a central point $\bp_0$ and prominent direction $\theta_n$, we remove that prominent direction from all candidate matches $\bp$ in the line segment. Multiple
crossing segments are naturally extracted by collecting position-orientation pairs in all prominent directions $\{\theta_n,1\leq n\leq N\}$. To enable the detection of crossing segments with very close direction angles, a fine discretization of the angle histogram is required. Alternatively, we can use variable bin sizes for each prominent direction, in which case only (a small length interval around) the angle of an extracted line would be removed from all candidate matches $\bp$ in the line segment. In the latter scenario, the corresponding length map may contain more than one peak, thus each one is dealt with independently.

\begin{figure}[htb]
 \centerline{\includegraphics[width=0.5\linewidth]{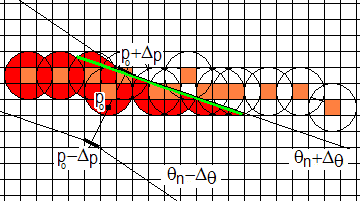}}
\caption{Edge points in a line outside the position-orientation range $[-\Delta_p,\Delta_p]\times [-\Delta_{\theta},\Delta_{\theta}]$ and a spurious line segment, shown in green, that could be erroneously detected inside that range.}\label{fig:linesOutsideIllustration}
\end{figure}

A final remark regards avoiding that a few edge points of lines with position-orientation parameters outside the range $[-\Delta_p,\Delta_p]\times [-\Delta_{\theta},\Delta_{\theta}]$ vote for spurious lines inside that range. If fact, this would happen whenever the uncertainty balls of those edge points intersect that range, as illustrated in Fig.~\ref{fig:linesOutsideIllustration}. We explicitly detect these cases and ignore them by using an orientation limit slightly larger than $\Delta_{\theta}$ (say, an increase of $2^{\circ}$) and only consider as detected segments those with estimated orientation within the original limits, {\it i.e.}, $\delta_{\theta}\in[-\Delta_{\theta},\Delta_{\theta}]$.

Alg.~\ref{alg:getFinalLine} synthesizes the procedure to extract line segments, where the usage of {\it non-maxima suppression} in line \ref{line:nms} is not detailed, since it is similar to the standard procedure for extracting peaks from the accumulator array of the HT. The only difference is that, since STRAIGHT has detected all the candidate matches for each line segment, the parameters $(\delta_p,\delta_{\theta})$ are more accurately estimated by fitting a line to the coordinates of the candidate matches with weights proportional to the magnitude of the image gradients. Naturally, other fitting criteria can easily be adopted in STRAIGHT.

\begin{algorithm}[hbt]
   \caption{Extracting line segments passing through $\bp_0$ with orientation close to $\theta_n$.\label{alg:getFinalLine}}
\begin{algorithmic}[1]
\STATE {\bf input:} $\bp_0$, $\theta_n$, half-plane length maps $\bL_{+}(\cdot,\cdot)$ and $\bL_{-}(\cdot,\cdot)$, prominent directions $\left\{\bTheta(\cdot)\right\}$, angle range limit $\Delta_{\theta}$, uncertainty ball radius $R$
\REPEAT
\STATE $(\delta_p,\delta_{\theta})=\arg\max \left[\bL_{+}(\cdot,\cdot)+\bL_{-}(\cdot,\cdot)\right]$ (find and remove peak using {\it non-maxima suppression}) \label{line:nms}
\IF{$|\delta_{\theta}| \leq \Delta_{\theta}$}
    \STATE $[\bp_+]=\mbox{round}\left(\bp_0+\bL_{+}(\delta_p,\delta_{\theta}) {\bv_{\theta_n+\delta_{\theta}}}/{\|\bv_{\theta_n+\delta_{\theta}}\|_{\infty}} +\delta_p \bv_{\theta_n+\delta_{\theta}}^{\bot}\right)$
    \STATE $[\bp_-]=\mbox{round}\left(\bp_0+\bL_{-}(\delta_p,\delta_{\theta}) {\bv_{\theta_n+\delta_{\theta}}}/{\|\bv_{\theta_n+\delta_{\theta}}\|_{\infty}} +\delta_p \bv_{\theta_n+\delta_{\theta}}^{\bot}\right)$
    \STATE for the {\it candidate matches} $\bp$ whose distance to $\overline{[\bp_-][\bp_+]}$ is smaller than (or equal to) $R$, remove from $\bTheta(\bp)$ the entry corresponding to $\theta_n$
    \ENDIF
\UNTIL there are not prominent peaks in $\left[\bL_{+}(\cdot,\cdot)+\bL_{-}(\cdot,\cdot)\right]$
\STATE {\bf output:} extremes of the extracted line segments, $\left\{[\bp_-],[\bp_+]\right\}$, and updated $\left\{\bTheta(\cdot)\right\}$
\end{algorithmic}
\end{algorithm}

\section{Hierarchical Implementation}
\label{sec:section4}

Although the computational complexity of the pixel-centered approach described in the previous section is much smaller than an intensive approach, there are still some issues that need addressing. Because the discretization of $\bL(\cdot,\cdot)$ must be fine, every time a new candidate match is found, a very large amount of positions in the length map need to be updated, which is a time-consuming operation. Furthermore, the number of pixels in the equidistant curves that fall within the search range and need to be scanned increases considerably with the size of the detected segment. Simultaneously, as the scanning of edge pixels proceeds and the corresponding updates are incorporated in the length map, the region of the map that remains updatable progressively becomes smaller. This occurs because more distant pixels correspond to smaller angle ranges, as explained in the previous section, and fewer $(\delta_p,\delta_{\theta})$ positions still correspond to quasi-connected line segments. Since this narrowing of the updatable area was not taken into account in the previous section, most pixel checks are unnecessary and further computational cost optimizations are possible. This motivates the hierarchical implementation of STRAIGHT, as outlined in this section.

\begin{figure}[htb]
\centerline{\includegraphics[width=0.4\linewidth]{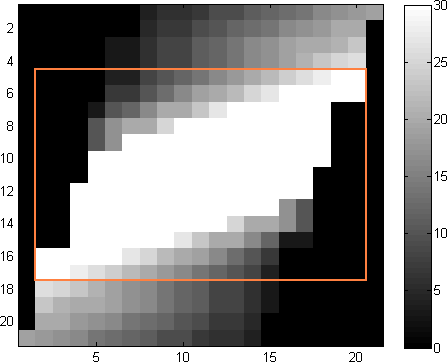}\hspace*{1cm}
\includegraphics[width=0.4\linewidth]{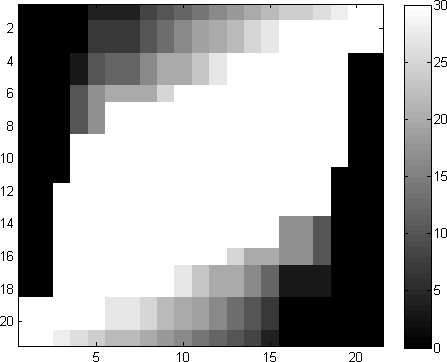}}
\caption{Illustration of the hierarchical implementation of STRAIGHT. Left: length map, with the bounding box of the updatable region. Right: the same region, after upscaling.}\label{fig:hierarchicalZoomingIn}
\end{figure}

Our hierarchical approach progressively zooms in on the updatable regions, thus increasing its discretization density. The process starts with a length map that spans the initial wide location and angle ranges, as described in the previous section. Every time a set of equidistant curves are processed, the rectangular bounding box containing the updatable region (illustrated in the left image of Fig.~\ref{fig:hierarchicalZoomingIn}) is upscaled, so that it takes up the complete length map (right image of Fig.~\ref{fig:hierarchicalZoomingIn}). This way, although the length map has a constant size, it progressively addresses narrower location and angle ranges around $(\delta_{p},\delta_{\theta})$, effectively increasing the resolution of the estimates. Since the resolution can increase indefinitely, a coarse discretization of $\bL(\cdot,\cdot)$ (we use an array of size $21\times 21$) becomes sufficient to obtain long line segments and fewer pixels are tested, thus resulting in computationally efficient line segment extractions. Since, as described in the previous section, a length map may contain multiple disconnected updatable regions, corresponding to different line segments passing through $\bp_0$, this process also divides the length map into multiple ones, each focused on a particular updatable region, and each estimation proceeds independently.

To implement the length map upscaling in the hierarchical STRAIGHT, we use Nearest Neighbor interpolation.

\section{Experiments}
\label{sec:section5}

In the absence of an established database for benchmarking the performance of methods for line segment extraction, we single out demonstrative results of STRAIGHT, contrasting them with the ones obtained with the HT~\cite{DudaHart72} and the state-of-the-art LSD~\cite{LSD10} (the superiority of LSD when compared to several other methods is thoroughly demonstrated in~\cite{LSD10}). We first describe experiments with synthetic images to illustrate extreme cases that help to characterize the general behavior of STRAIGHT. Then, we present results obtained with several real world images that demonstrate its performance in practice.

\subsection{Synthetic images}

\begin{figure}[htb]
\centerline{
\includegraphics[width=.25\linewidth]{synthetic2small.PNG}
\includegraphics[width=.25\linewidth]{synthetic2smallHoughResult.PNG}
\includegraphics[width=.25\linewidth]{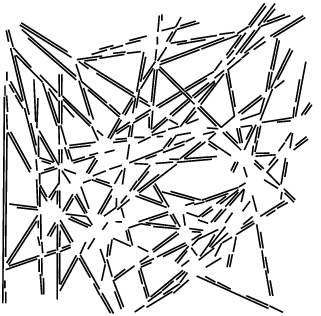}
\includegraphics[width=.25\linewidth]{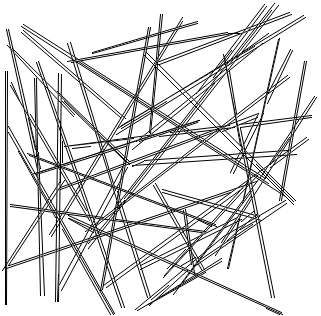}}
\caption{Clutterless image with prominent lines. From left to right: original binary image, result of the HT \cite{DudaHart72}, LSD \cite{LSD10}, and the proposed method STRAIGHT.}\label{fig:syntheticImages}
\end{figure}

We start by illustrating that STRAIGHT succeeds in cases tailored to the HT, {\it i.e.}, when processing images for which the HT exhibits clear superiority with respect to local methods. We use again the synthetic image used in the Section~\ref{sec:intro}, reproduced on the left of Fig.~\ref{fig:syntheticImages}. This is a binary image used in a review of several HT-based line segment extraction methods \cite{probalisticAndNonProbabilistic95}. As we have anticipated in the Section~\ref{sec:intro}, and in accordance with the conclusions of \cite{probalisticAndNonProbabilistic95}, the HT succeeds in correctly extracting the lines from this image. In fact, although the multiple crossings make this image visually complex, the HT accumulator array exhibits the desired prominent peaks (see Fig.~\ref{fig:houghIllustration}), capturing the fact that the lines are long and not in a very large number. In the third image of Fig.~\ref{fig:syntheticImages}, we show the results of LSD. That a pair of twin segments is extracted for each one in the original image is due to the fact that LSD treats the binary image as any other, {\it i.e.}, as a grey-level one, and both light-to-dark and dark-to-light transitions are detected. However, what is more important is that the local nature of the LSD limits its performance, particularly in resolving the line intersections, making it fail the extraction of several complete segments that cross each other. The rightmost image displays the result of STRAIGHT, showing that it successfully extracts the majority of the line segments, regardless of the intersections (to make the comparison fair, we also processed the image as a grey-level one, originating the double-detection effect).

\begin{figure}[htb]
\centerline{
\includegraphics[width=.25\linewidth]{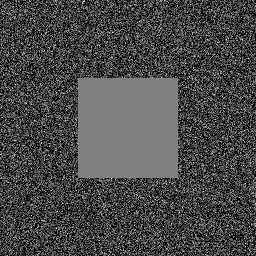}
\includegraphics[width=.25\linewidth]{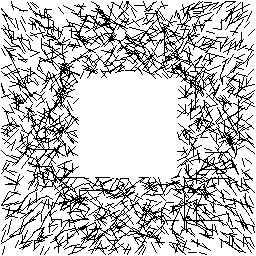}
\includegraphics[width=.25\linewidth]{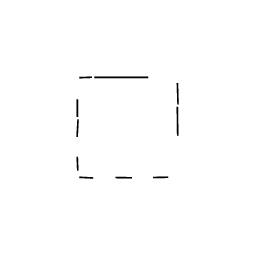}
\includegraphics[width=.25\linewidth]{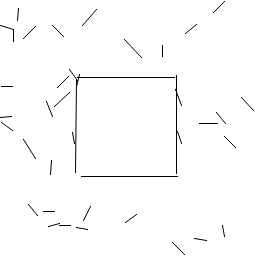}
}
\vspace{0.25cm}
\centerline{
\includegraphics[width=.25\linewidth]{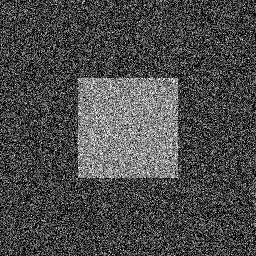}
\includegraphics[width=.25\linewidth]{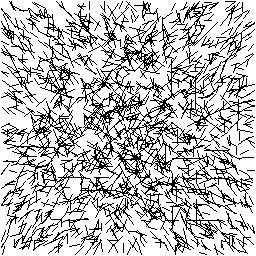}
\includegraphics[width=.25\linewidth]{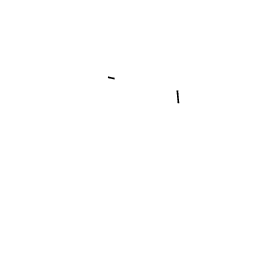}
\includegraphics[width=.25\linewidth]{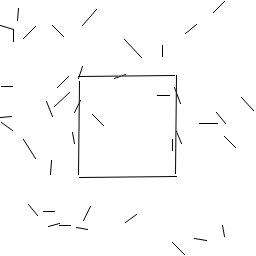}
}
\caption{Textured images. From left to right: original image, result of the HT \cite{DudaHart72}, LSD \cite{LSD10}, and STRAIGHT.}\label{fig:noiseResults}
\end{figure}

We now illustrate the behavior of the algorithms when dealing with the other extreme of the spectrum, {\it i.e.}, with images whose line segments are characterized by being frontiers of differently textured regions, rather than abrupt changes in a very smooth intensity level. We use the synthetic images in the left of Fig.~\ref{fig:noiseResults}, which were generated by adding noise to a piecewise constant map. The top image simulates a scenario where a textureless objects occludes a textured one ({\it e.g.}, a wall in front of a tree) and the bottom one simulates two textured objects. In both cases, although the line segments that separate regions are perceptually evident, they are not trivially mapped to the edges of the images. In fact, since the textures produce a large number of spurious edge points and the perceptual segments do not guaranteedly produce the corresponding edges, the HT~\cite{DudaHart72} fails to extract them and originates a huge number of false detections, as shown in Fig.~\ref{fig:noiseResults}. Differently, LSD~\cite{LSD10} succeeds in interpreting the textures as not forming line segments but only captures parts of the real segments for the top image and almost none for the bottom one. This is due to the local nature of LSD, which makes it sensitive to the missing edge points in the line segments. The rightmost images of Fig.~\ref{fig:noiseResults} display the results of STRAIGHT, showing that it succeeds in extracting the perceptually relevant lines as forming, in both cases, four complete line segments (the few short segments correspond to accidental connected alignments in the random texture).

\subsection{Real images}

\begin{figure}[htb!]
\centerline{
\includegraphics[width=.45\linewidth]{Buildings.jpg}\hspace*{.5cm}
\includegraphics[width=.45\linewidth]{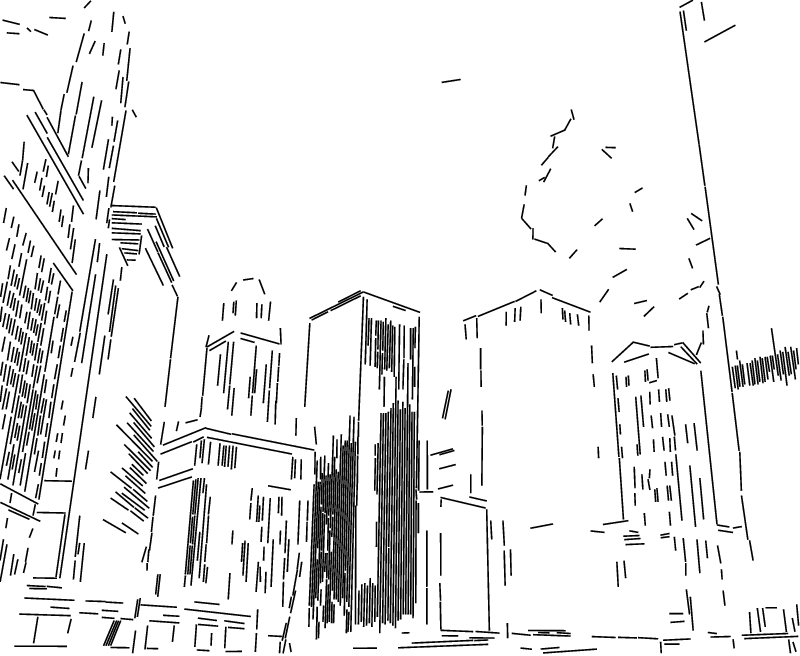}
}\vspace*{.5cm}
\centerline{
\includegraphics[width=.45\linewidth]{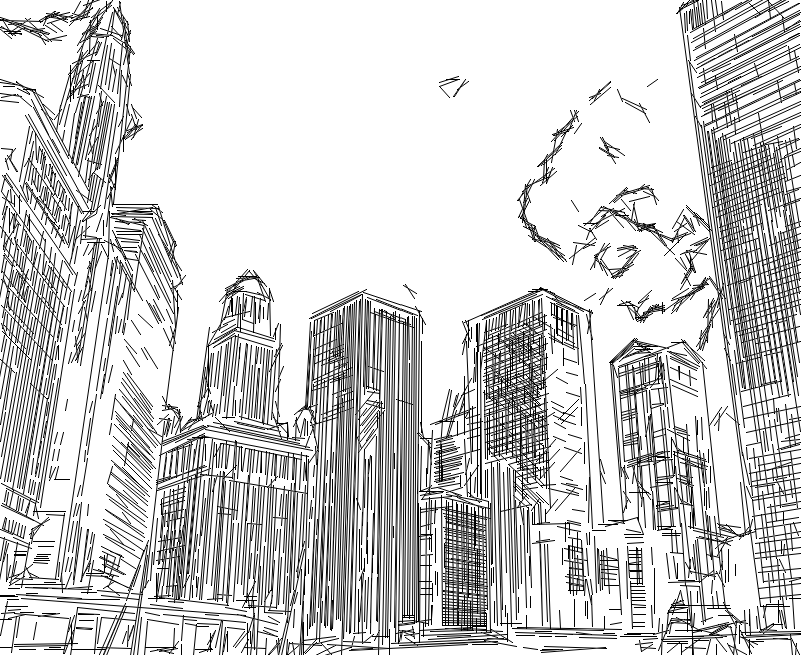}\hspace*{.5cm}
\includegraphics[width=.45\linewidth]{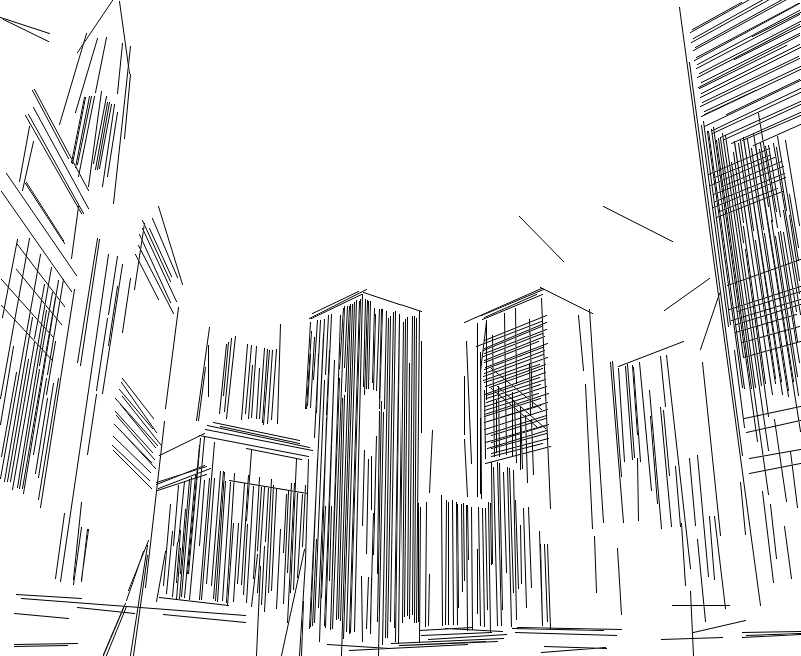}
}
\caption{Top left: image. Top right: LSD~\cite{LSD10}. Bottom left: STRAIGHT. Bottom right: STRAIGHT (longer line segments).}\label{fig:building}
\end{figure}

We start be showing the results obtained with the image used in Section~\ref{sec:intro} to clarify the limitations of the HT (Fig.~\ref{fig:houghIllustration}). This image is challenging due to its dense packing of line segments of multiple lengths. In the top right image of Fig.~\ref{fig:building}, we display the results of LSD~\cite{LSD10}, showing that a subset of the line segments are in fact detected. However, a closer look reveals that those are only the line segments that do not cross other structures and also that several longer segments are detected as fragmented ones. The results of STRAIGHT are in the two bottom images of Fig.~\ref{fig:building}. We see that our method succeeds in extracting the vast majority of the line segments in the image (exceptions are those which exhibit very low contrast). The fact that the extracted line segments are complete is particularly evident in the bottom right image, which displays only the line segments that have length greater than 50 pixels.

\begin{figure}[htb]
\centerline{
\includegraphics[width=.33\linewidth]{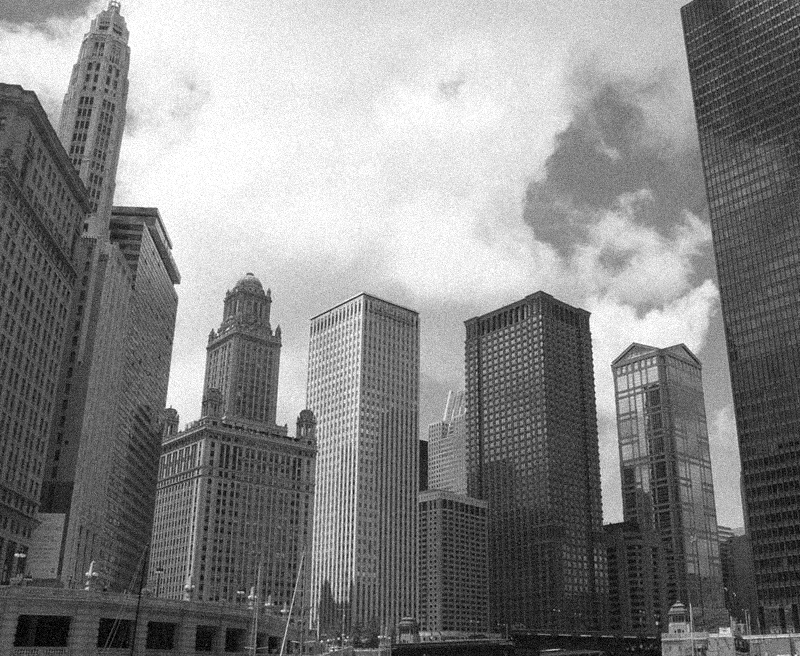}
\includegraphics[width=.33\linewidth]{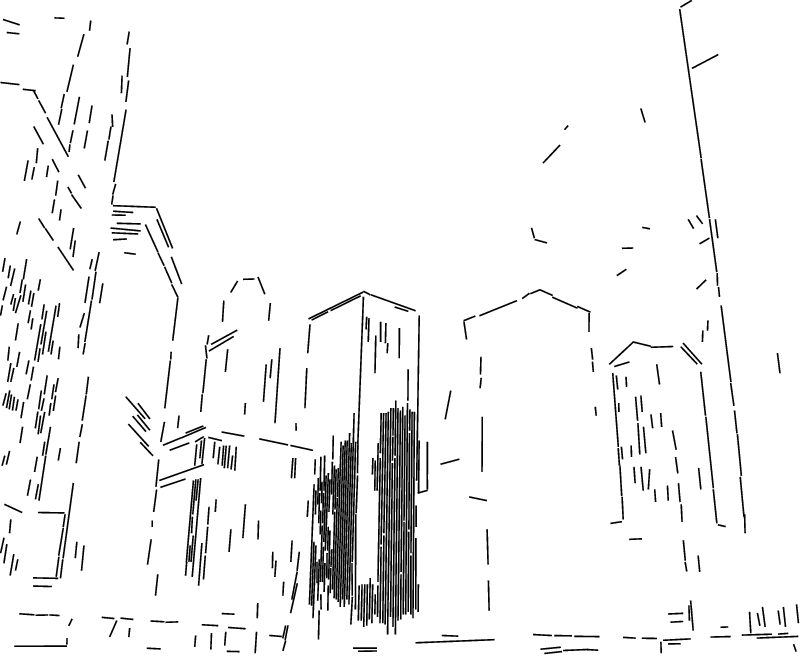}
\includegraphics[width=.33\linewidth]{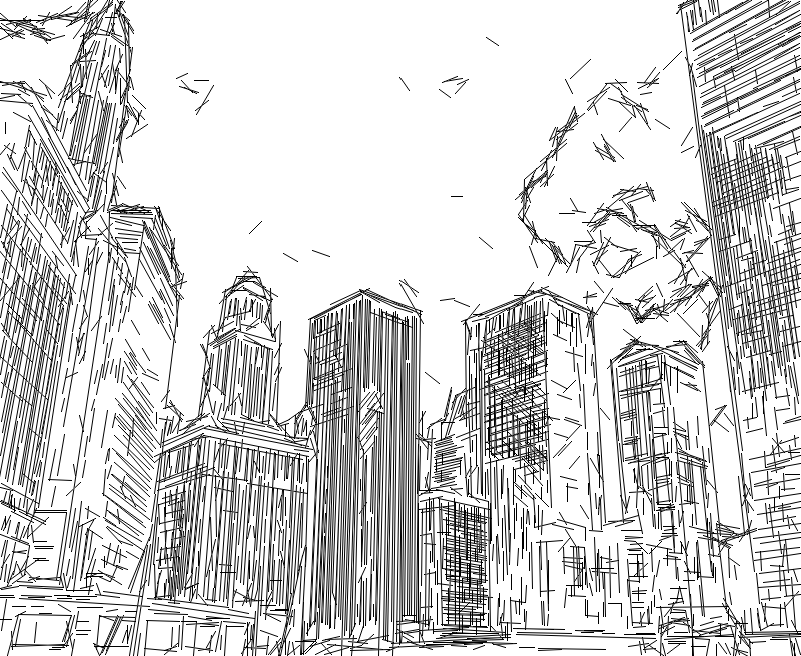}}\vspace*{.25cm}
\centerline{
\includegraphics[width=.33\linewidth]{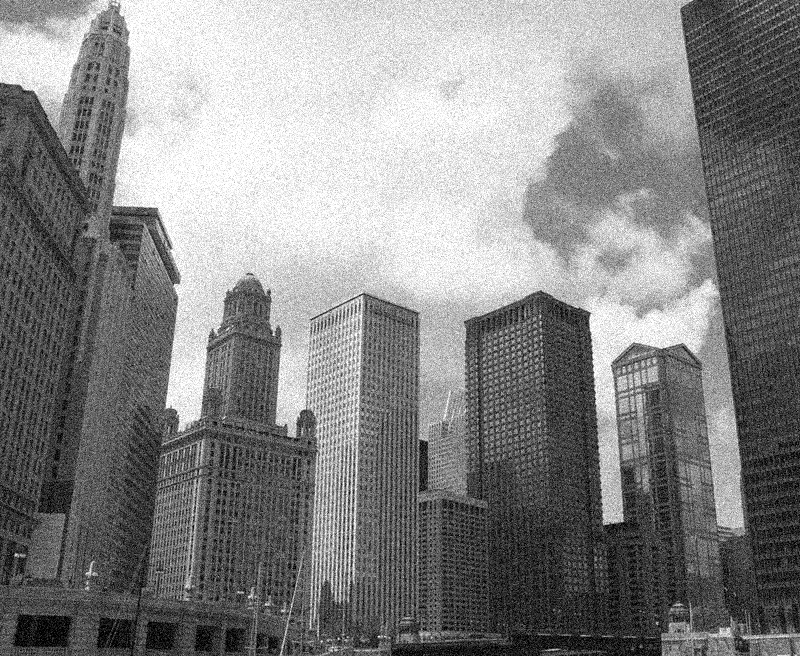}
\includegraphics[width=.33\linewidth]{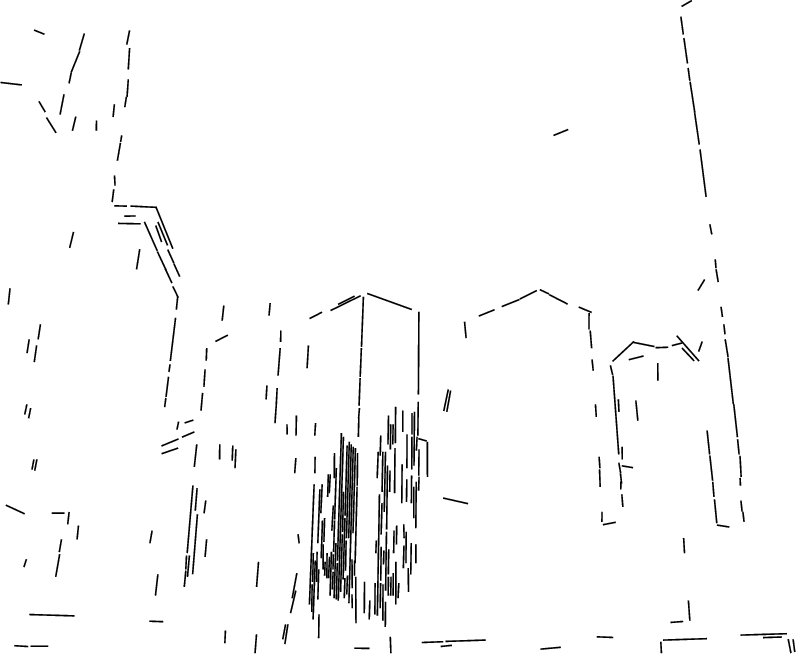}
\includegraphics[width=.33\linewidth]{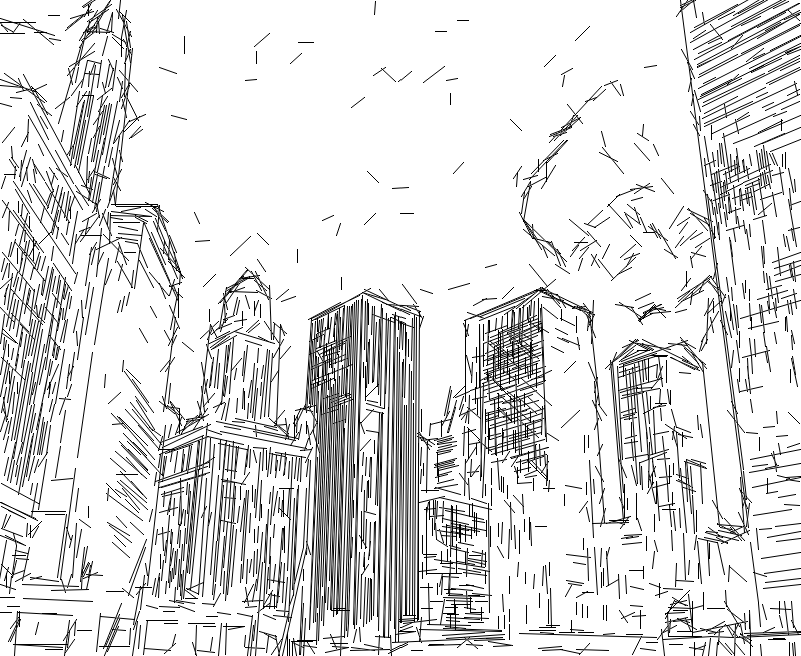}
}
\caption{Left: noisy images ($\sigma \!=\! 10$ on the top and $\sigma \!=\! 20$ on the bottom). Center: LSD~\cite{LSD10}. Right: STRAIGHT.}\label{fig:noisyBuildings}
\end{figure}

To illustrate how the noise affects the extraction of line segments in real images, we report the results obtained with noisy versions of the same image. Fig.~\ref{fig:noisyBuildings} synthesizes the results for two levels of zero-mean white Gaussian noise. We see that, with the increase of the noise level, LSD~\cite{LSD10} originates more segment fragmentations and a progressive failure to detect some line segments. The performance of STRAIGHT declines in a less steep way, as expected from its global nature.

\begin{figure}[htb]
\centerline{
\includegraphics[width=.45\linewidth]{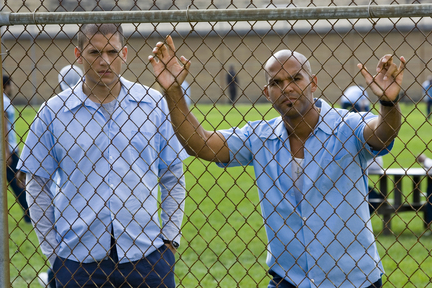}\hspace*{.5cm}
\includegraphics[width=.45\linewidth]{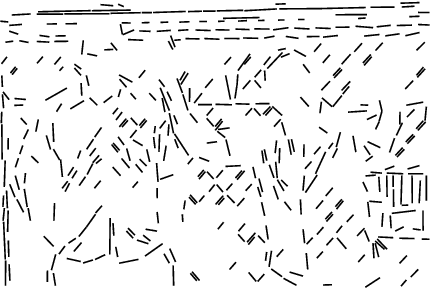}
}\vspace*{.5cm}
\centerline{
\includegraphics[width=.45\linewidth]{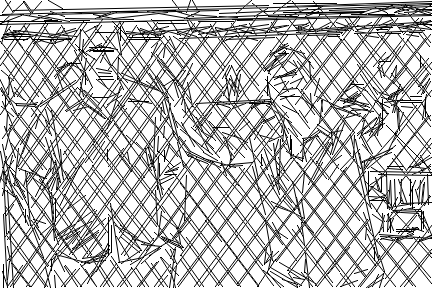}\hspace*{.5cm}
\includegraphics[width=.45\linewidth]{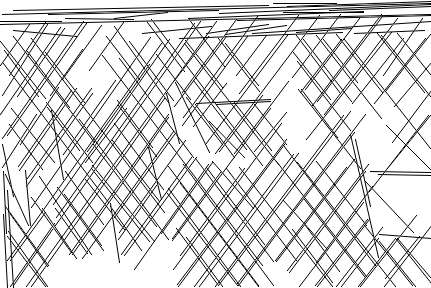}
}
\caption{Top left: image. Top right: LSD~\cite{LSD10}. Bottom left: STRAIGHT. Bottom right: STRAIGHT (longer line segments).}\label{fig:prisonBreak}
\end{figure}

Fig.~\ref{fig:prisonBreak} presents another illustrative case. It was obtained by processing an image containing a complex scene occluded by a net composed of very long line segments that cross multiple times. The result of  LSD~\cite{LSD10} shows the net broken into short line segments (several sections of the net are not even extracted). On the other hand, our method was able to obtain almost all the complete line segments of the net, even in locations where the background is complex (exceptions are where the net has a very low contrast with respect to the background). The line segments extracted by our method that have length greater than 50 pixels, displayed in the bottom right image of Fig.~\ref{fig:prisonBreak}, make this particularly evident.

\begin{figure}[htb]
\centerline{
\includegraphics[width=.25\linewidth]{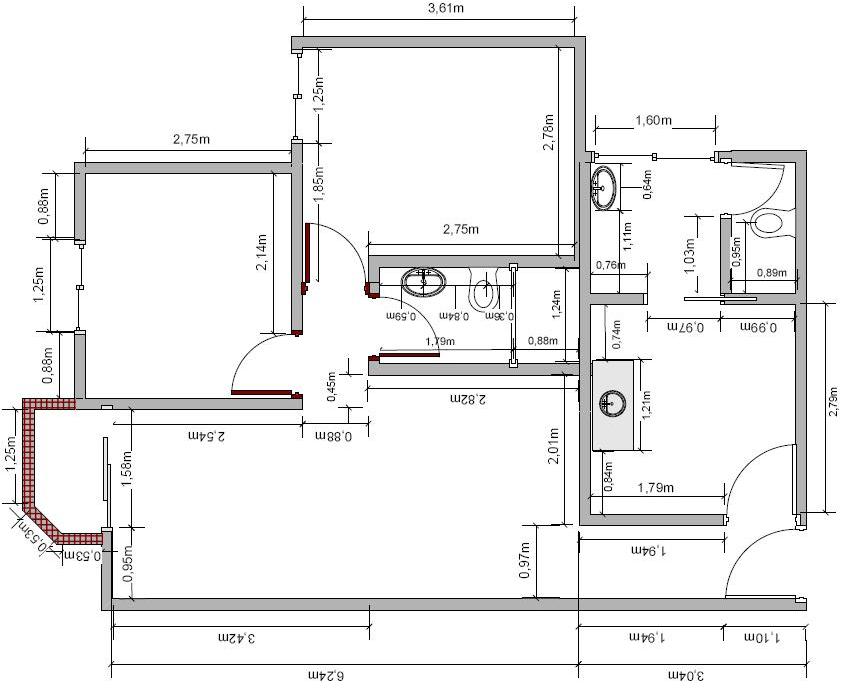}
\includegraphics[width=.25\linewidth]{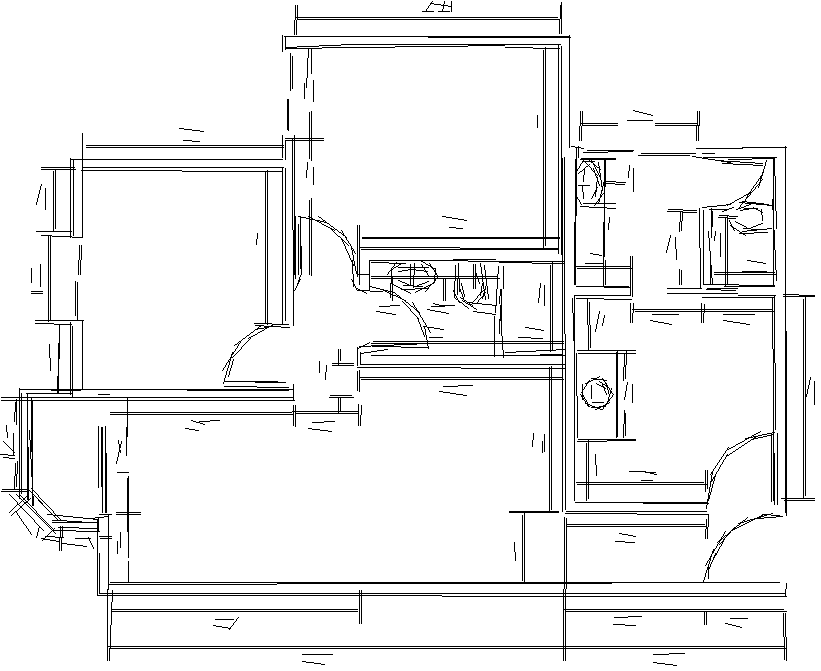}
\includegraphics[width=.25\linewidth]{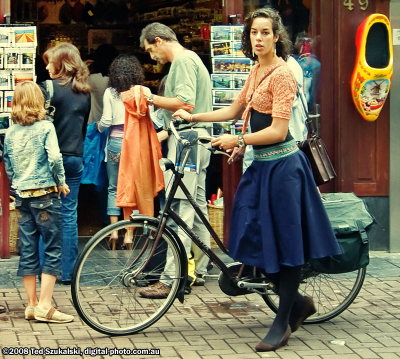}
\includegraphics[width=.25\linewidth]{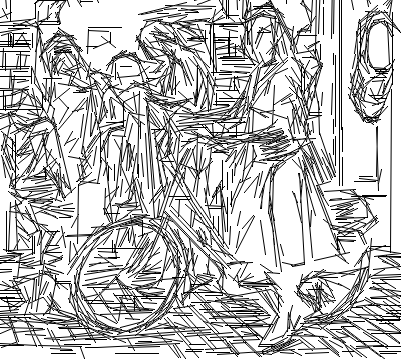}
}\vspace*{.1cm}
\centerline{
\includegraphics[width=.25\linewidth]{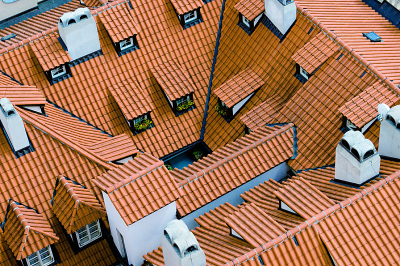}
\includegraphics[width=.25\linewidth]{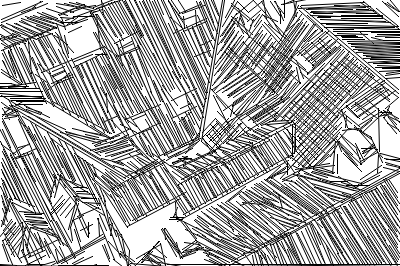}
\includegraphics[width=.25\linewidth]{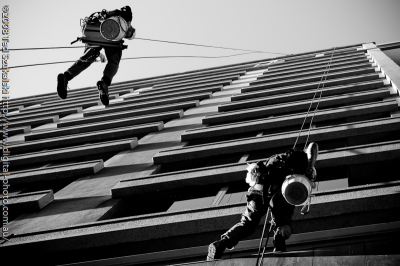}
\includegraphics[width=.25\linewidth]{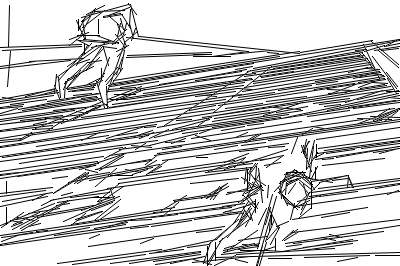}
}\vspace*{.1cm}
\centerline{
\includegraphics[width=.25\linewidth]{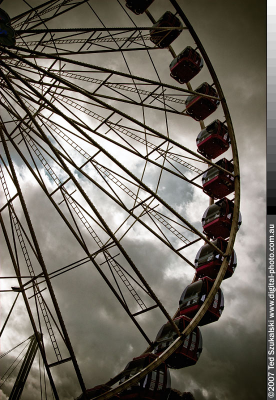}
\includegraphics[width=.25\linewidth]{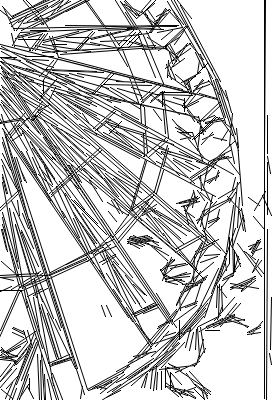}
\includegraphics[width=.25\linewidth]{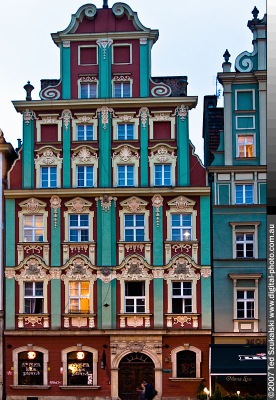}
\includegraphics[width=.25\linewidth]{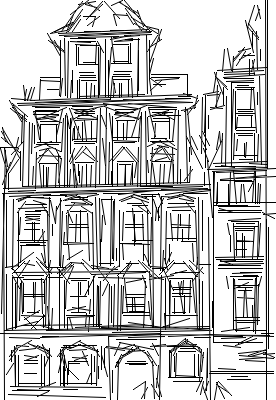}
}\vspace*{.1cm}
\centerline{
\includegraphics[width=.25\linewidth]{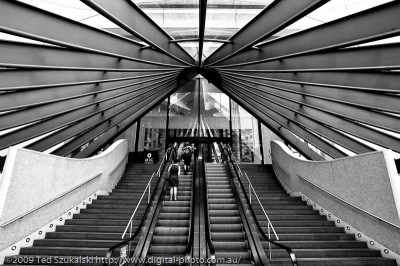}
\includegraphics[width=.25\linewidth]{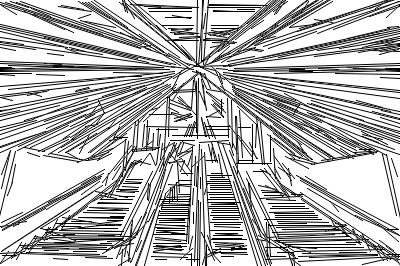}
\includegraphics[width=.25\linewidth]{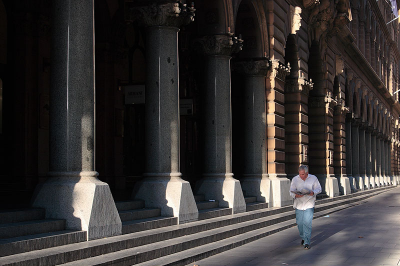}
\includegraphics[width=.25\linewidth]{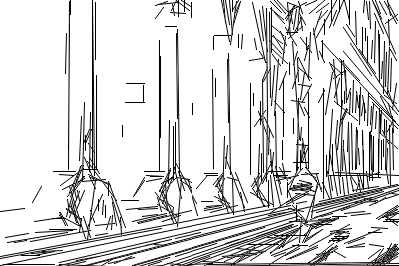}
}\vspace*{.1cm}
\centerline{
\includegraphics[width=.25\linewidth]{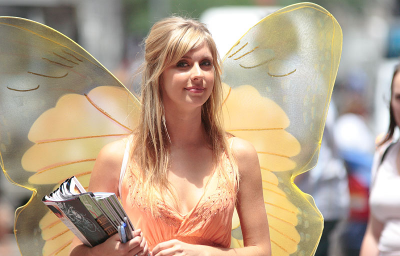}
\includegraphics[width=.25\linewidth]{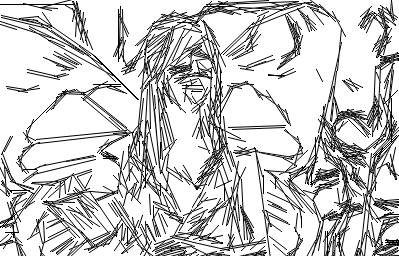}
\includegraphics[width=.25\linewidth]{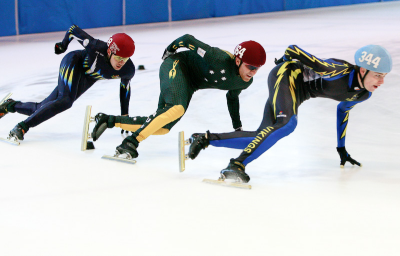}
\includegraphics[width=.25\linewidth]{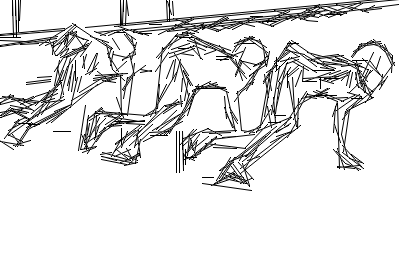}
}\vspace*{.1cm}
\centerline{
\includegraphics[width=.25\linewidth]{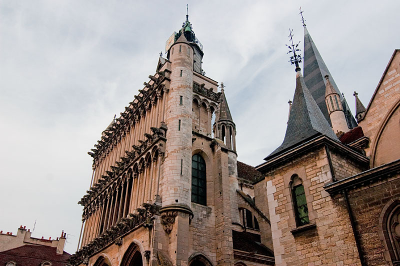}
\includegraphics[width=.25\linewidth]{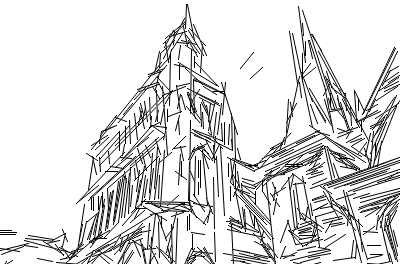}
\includegraphics[width=.25\linewidth]{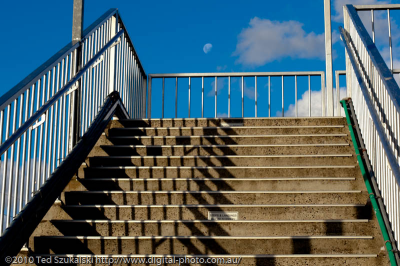}
\includegraphics[width=.25\linewidth]{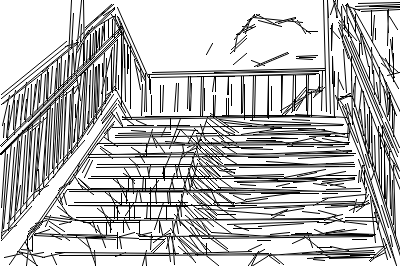}
}
\caption{Results of STRAIGHT for several kinds of real images.}\label{fig:manyResults}
\end{figure}

Finally, Fig.~\ref{fig:manyResults} presents results of using STRAIGHT with real images of various kinds. As desired, the vast majority of long line segments are extracted without artificial fragmentation, despite the multiple segment crossings. Also note that, although some of these images have edges that form curves, STRAIGHT succeeds in approximating these sections in a piecewise linear way, {\it i.e.}, by a sequence of rectilinear line segments.

\section{Conclusion}
\label{sec:conclusions}

We have presented a new method for line segment extraction, which we call STRAIGHT (Segment exTRAction by connectivity-enforcIng Hough Transform). Our method inherits the global accuracy of the HT and overcomes its limitations, particularly those that arise from not taking into account that line segments are {\it connected} sets of edge points. Our experiments show that STRAIGHT outperforms current methods for line segment extraction in challenging situations, {\it e.g.}, when dealing with complex images containing several crossing segments.

We end by pointing out that our approach may pave the way to other improvements in HT-like image edge analysis. In fact, as we saw, the HT leads to erroneous votes, which are eliminated by taking point connectivity into account. Thus, the detection of non-rectilinear shapes, {\it e.g.}, circles, in challenging scenarios, may also benefit from a similar treatment.

\section*{Acknowledgements}

This work was partially supported by FCT, under ISR/IST plurianual funding, through the PIDDAC Program, and grants MODI-PTDC/EEA-ACR/72201/2006 and SFRH/BD/48602/2008.

\bibliographystyle{IEEEtran}

\bibliography{IEEEfull,references}

\end{document}